\documentclass[runningheads]{llncs}
\usepackage[T1]{fontenc}
\usepackage{graphicx}
\usepackage{booktabs}
\usepackage[misc]{ifsym}
\newcommand{\corr}{(\Letter)}
\usepackage{amsmath,amssymb}

\usepackage{float}

\renewcommand{\thefigure}{S\arabic{figure}}
\renewcommand{\thetable}{S\arabic{table}}

\newcommand{\ndstarBNsmall}{0.93}
\newcommand{\ndstarBNfive}{0.89}
\newcommand{\ndstarBNten}{0.84}
\newcommand{\ndstarBNlarge}{0.75}
\newcommand{\ndstarMonotone}{monotone}
\newcommand{\nhTwoAdvRhoZero}{0.34}
\newcommand{\nhTwoAdvRhoHi}{3.47}
\newcommand{\nhTwoTten}{2.73}
\newcommand{\nhTwoTfifty}{3.47}
\newcommand{\nhTwoTtwohundred}{3.72}
\newcommand{\nhTwoPhiZero}{0.29}
\newcommand{\nhTwoPhiOne}{3.47}
\newcommand{\nepsCrossDivLo}{14.27}
\newcommand{\nepsCrossConfLo}{14.04}
\newcommand{\nepsCrossDivHi}{13.56}
\newcommand{\nepsCrossConfHi}{14.26}
\newcommand{\nclusterLo}{7.59}
\newcommand{\nclusterHi}{9.31}
\newcommand{\nbetaShockAdvHi}{2.94}
\newcommand{\nwAdvRho}{3.47}
\newcommand{\nwAdvRhoSq}{3.46}
\newcommand{\nwAdvRhoHalf}{3.47}
\newcommand{\nvacRnone}{17.5}
\newcommand{\nvacTau}{0.1}
\newcommand{\nvacNa}{10}
\newcommand{\nvacNb}{6}
\newcommand{\nfleetRhoZeroGsm}{0.52}

\newcommand{\nnBeyondGsm}{4}

\newcommand{\ngptAurocGsm}{0.59}
\newcommand{\ngptDeltaPrecGsm}{0.68}
\newcommand{\ngptDeltaStarGsm}{0.83}

\newcommand{\nfleetRhoZeroHpqa}{0.70}

\newcommand{\nnBeyondHpqa}{2}

\newcommand{\ngptAurocHpqa}{0.56}
\newcommand{\ngptDeltaPrecHpqa}{0.75}

\newcommand{\neceMin}{0.43}
\newcommand{\neceMax}{0.94}

\newcommand{\ngptHpqaSmallPrec}{0.870}
\newcommand{\ngptHpqaSmallStar}{0.887}
\newcommand{\ndstarMatchedGsmLo}{0.83}

\newcommand{\neTwoRowsBoth}{Qwen3-0.6B & 0.94 & 0.50 & 0.84 & 0.94 & 0.77 & 0.50 & 0.83 & 0.76 \\
Qwen3-4B & 0.76 & 0.50 & 0.88 & 0.76 & 0.46 & 0.51 & 0.86 & 0.46 \\
Qwen3-8B & 0.73 & 0.50 & 0.86 & 0.73 & 0.43 & 0.50 & 0.81 & 0.43 \\
Mistral-7B-Instr. & 0.92 & 0.51 & 0.81 & 0.92 & 0.52 & 0.52 & 0.78 & 0.51 \\
Phi-4-mini & 0.79 & 0.51 & 0.87 & 0.78 & 0.60 & 0.50 & 0.86 & 0.59 \\
gpt-4o-mini & 0.64 & 0.59 & 0.68 & 0.63 & 0.40 & 0.56 & 0.75 & 0.38 \\}
\newcommand{\nparseFailMax}{3.2}
\newcommand{\nvarCmaxOpen}{0.016}
\newcommand{\neHatOpenLo}{0.43}
\newcommand{\neHatOpenHi}{0.94}
\newcommand{\naurocOpenLo}{0.50}
\newcommand{\naurocOpenHi}{0.52}
\newcommand{\naurocAtFlip}{0.34}
\newcommand{\nreplayImputedPct}{0.6}
\newcommand{\nrhoAdjGsm}{0.72}
\newcommand{\nrhoAdjHpqa}{0.93}
\newcommand{\nrhoSmallLo}{0.39}
\newcommand{\nrhoSmallHi}{0.61}
\newcommand{\nrhoInterLo}{0.33}
\newcommand{\nrhoInterHi}{0.85}
\newcommand{\nrhoInterIndLo}{0.14}
\newcommand{\nrhoInterIndHi}{0.63}
\newcommand{\nrhoGptLo}{0.27}
\newcommand{\nrhoGptHi}{0.88}

\newcommand{\neThreeRandomGsm}{10.89}
\newcommand{\neThreeConfGsm}{10.85}
\newcommand{\neThreeDivGsm}{10.75}
\newcommand{\neThreeOracleGsm}{10.58}

\newcommand{\nholmRandDivGsmP}{\ensuremath{\le 0.002}}
\newcommand{\neThreeRandDivGsmMean}{0.133}
\newcommand{\nholmConfDivGsmP}{\ensuremath{\le 0.002}}
\newcommand{\neThreeConfDivGsmMean}{0.095}
\newcommand{\nholmRandDivHpqaP}{\ensuremath{\le 0.002}}
\newcommand{\neThreeRandDivHpqaMean}{0.322}
\newcommand{\nholmConfDivHpqaP}{\ensuremath{\le 0.002}}
\newcommand{\neThreeConfDivHpqaMean}{0.312}
\newcommand{\ndRhoGsm}{-0.009}
\newcommand{\ndRhoGsmLo}{-0.064}
\newcommand{\ndRhoGsmHi}{0.047}

\newcommand{\ndRhoHpqa}{0.023}
\newcommand{\ndRhoHpqaLo}{-0.035}
\newcommand{\ndRhoHpqaHi}{0.086}

\begin{document}

\title{One Human, $N$ Agents: Audit-Budget Allocation for LLM Agent Fleets
under Miscalibrated, Correlated Confidence}
\titlerunning{One Human, $N$ Agents}

% Double-blind submission (per the ECML PKDD 2026 author kit): author
% information scrubbed; \authorrunning commented out.
%\author{Author information scrubbed for double-blind reviewing}
%\institute{Institute information scrubbed for double-blind reviewing}
%
% --- Camera-Ready Version (uncomment after acceptance, delete the two lines above) ---
\author{Cesare Zavattari  \and Alessandro Tommasi \and Giuseppe Prencipe \corr}
\authorrunning{C. Zavattari et al.}
\institute{Dipartimento di Informatica, Universit\`a di Pisa, Pisa, Italy}

\maketitle

\begin{abstract}
A single human must audit $N$ LLM agents under a budget of $B \ll N$ audits
per round, guided by self-reported confidence that may be adversarially
miscalibrated and by correlated errors. We model this as budgeted noisy
inspection over a two-level Gaussian copula and locate the miscalibration
threshold $\delta^*$ past which confidence-ranked auditing is \emph{worse}
than random. Two a-priori expectations reverse: $\delta^*$ \emph{rises}
as the budget shrinks, and cross-family correlation is not low---shared
difficulty dominates lineage. Five open-weight LLMs show operationally useless
(near-constant) confidence, point estimates at or beyond the flip though
CIs straddle it; a proprietary model is informative and lands below it. We give a
quantitative criterion for \emph{vacuous} oversight, and replaying policies
on recorded traces confirms the ordering.

\end{abstract}

\section{Introduction}\label{sec:intro}

Deployments increasingly run not one LLM agent but a \emph{fleet}: $N$
persistent agents built from a few base models, emitting outputs faster
than any human can check. Regulation nonetheless mandates
\emph{effective} human oversight of high-risk systems (EU AI Act,
Art.~14 \cite{euaiact2024}), while one supervisor can audit $B \ll N$
outputs per round. Auditing ``the least confident outputs'' trusts a signal
supplied by the auditees themselves---and though LLMs mostly know what they
know \cite{kadavath2022}, their verbalized confidence is badly calibrated
\cite{lin2022teachingmodelsexpressuncertainty,tian2023,xiong2024}. Throughout, agents are
\emph{honest-but-miscalibrated}; strategic behavior is the province of AI
Control \cite{greenblatt2024} and future work.

We formalize budgeted noisy inspection of a fleet with two coupled
imperfections: self-reported confidence whose informativeness degrades with
an adversarial-miscalibration strength $\delta$, and errors correlated
through a two-level Gaussian copula. We ask when confidence-ranked auditing
flips to \emph{worse} than random (threshold $\delta^*$); how much
correlation-aware posterior transfer helps; and when oversight is
\emph{vacuous}---no feasible policy beats no oversight by more than a
fraction $\tau$, a quantitative reading of ``rubber-stamping''
\cite{green2022}.

To the best of our knowledge, no prior work jointly models (i) a single
human supervisor allocating (ii) a limited audit budget across (iii) a fleet
of persistent LLM agents whose (iv) self-reported confidence is subject to
adversarial miscalibration and whose (v) error correlation is partially
predictable from fleet composition (capability tier and lineage); nor does
prior work connect such allocation to a quantitative criterion for vacuous
human oversight.

We measured the premises before trusting the phase diagram, and two
a-priori expectations reversed: $\delta^*$ \emph{rises} as the budget
shrinks (\S\ref{sec:synthetic}), and shared item difficulty, not lineage,
dominates cross-family correlation (\S\ref{sec:measurement}).
\emph{Contributions}: the model with its located flip threshold
$\delta^*(B/N)$ and vacuity criterion (\S\ref{sec:model}--\ref{sec:synthetic});
locked measurements of six LLMs spanning the spectrum; a trace-grounded
replay confirming the ordering, all pre-specified contrasts
Holm-signi\-ficant (\S\ref{sec:measurement}).

Illustrative material for the paper is collected in the Appendix.

\section{Related Work}\label{sec:related}

\paragraph{Scoping: deferral allocates instances; we allocate audits.}
Selective prediction decides, per \emph{instance}, whether the model predicts
or abstains \cite{pugnana2023}; learning-to-defer instead routes the instance
to a model or a human \cite{mozannar2020,palomba2025}; multi-expert L2D routes
instances to experts
\cite{verma2023,mao2023two,mao2023principled,alves2024}. They allocate
\emph{instances to experts}; we allocate \emph{audits to persistent
agents}: the human never replaces an output, only spends a budget to catch
errors. Five features separate this from generic budgeted selection:
(1)~the units are persistent agents, not data points; (2)~their group
structure is given by
the fleet's bill of materials, not estimated; (3)~error correlation is
partially predictable from fleet composition (chiefly capability tier;
\S\ref{sec:measurement}); (4)~an audit on one agent updates beliefs about
its siblings; and (5)~the priority signal is supplied by the audited agent
itself, not estimated by the allocator---which---once adversarially
miscalibrated---is exactly what makes confidence-ranked auditing flip to
worse-than-random (\S\ref{sec:synthetic}).

The problem touches five literatures. \emph{Budgeted inspection}: sequential
search goes back to Pandora's box \cite{weitzman1979}, acceptance sampling
\cite{dodgeromig1959}, and skip-lot schemes (ISO 2859-3 \cite{iso2859}) even
modulate inspection by supplier self-reports, without a formal treatment of
adversarial miscalibration or cross-unit correlation. \emph{Budgeted query
allocation in ML}: active learning and crowd labeling allocate queries under
budgets \cite{hacohen2022,chen2013}, batch multi-fidelity methods decorrelate
batches \cite{li2022bmfal}, and correlated bandits exploit known arm
correlation \cite{gupta2021correlated,gupta2020latent}---the closest
mechanism to ours, but they maximize reward from signals the \emph{learner}
observes, whereas we detect errors where the priority signal is supplied
\emph{by the auditee}. \emph{AI control} protocols audit untrusted models
assuming \emph{strategic} subversion \cite{greenblatt2024}, leaving the
honest-but-miscalibrated fleet with a human allocator open. On
\emph{confidence}, LLMs mostly know what they know \cite{kadavath2022}, yet
their verbalized estimates are systematically overconfident
\cite{lin2022teachingmodelsexpressuncertainty,tian2023,xiong2024}, worse in agentic settings
\cite{confdichotomy2026,agenticcal2026}; we consume these as premises and
measure their allocation-level consequence. Finally, \emph{socio-legal}
oversight mandates are criticized as rubber-stamping \cite{green2022}, with
fleet-level accountability gaps \cite{euaiact2024,butt2026}; for vacuous
oversight we found no explicit quantitative criterion---H3 supplies one.

\section{Model}\label{sec:model}

The lineage is acceptance sampling \cite{dodgeromig1959} extended with
priority signals that are self-reported, adversarially miscalibrated, and
correlated across units.

\paragraph{Fleet and errors.} $N$ persistent agents are partitioned into $F$
base-model families. In round $t$, agent $i$ errs with indicator
$E_i \in \{0,1\}$, $\Pr(E_i{=}1) = e_i \sim U(0.05,0.30)$ fixed per agent;
correlation enters through a two-level Gaussian copula
$Z_i = \sqrt{\rho_0}\, G_0 + \sqrt{\rho_f - \rho_0}\, G_{\mathrm{fam}(i)} +
\sqrt{1-\rho_f}\, U_i$, with $E_i = \mathbf{1}\{\Phi(Z_i) < e_i\}$ and
$G_0, G_f, U_i \sim \mathcal N(0,1)$: $\rho_0$ a fleet-wide shared factor
(e.g.\ item difficulty), $\rho_f \ge \rho_0$ the within-family latent
correlation. The shocks are drawn \emph{once per replication} and evolve as
stationary AR(1) with persistence $\varphi$ (default 1): at temperature~0,
same-family co-failure is a stable difficulty profile, not per-round
noise---persistence makes an audit on one agent evidence about its siblings.
Marginals are unaffected. $\rho$ is quoted in \emph{latent} units;
estimating $\hat\rho$ inverts the copula map given measured marginals
(caveat in \S\ref{sec:limitations}).

\paragraph{Confidence.} Correct outputs report confidence
$c \sim \mathrm{Beta}(6,2)$; errors draw from the mixture
$(1-\delta)\,\mathrm{Beta}(3,3) + \delta\,\mathrm{Beta}(10,1.5)$: $\delta$
is the fraction of errors whose confidence mimics that of correct outputs.

\paragraph{Audit and risk.} Per round the supervisor audits $A_t$,
$|A_t| \le B$; an audit reveals $E_i$ flipped with probability $\varepsilon$.
The metric is \emph{residual risk} $R(\pi) = \mathbb E[\sum_i E_i \mathbf
1\{i \notin \mathrm{caught}(A_t)\}]$, the expected undetected errors per
round. Oversight is \emph{vacuous} at level $\tau$ if $\max_\pi
(R_{\mathrm{none}} - R_\pi)/R_{\mathrm{none}} < \tau$ over the deployable
policies considered.

\paragraph{Policies.} \textsf{random}; \textsf{conf\_ranked} ($B$ lowest,
ties random); \textsf{round\_robin}; \textsf{oracle\_conf} (confidence at
$\delta{=}0$: upper reference, not deployable); \textsf{diversity\_bayes}:
per-agent Beta posteriors, audits update same-family siblings with weight $w$
(robust over $\{\rho,\rho^2,\rho/2\}$), batched-greedy posterior UCB---a
knowledge-gradient-style instantiation \cite{chen2013,gupta2021correlated},
not an optimality claim.

\section{Synthetic Phase Diagram (E1)}\label{sec:synthetic}

We sweep $N \in \{20,100\}$, $F \in \{1,4\}$,
$B/N \in \{0.02,\dots,0.2\}$, $\delta \in \{0,\dots,1\}$,
$\rho \in \{0,\dots,0.8\}$, $\varepsilon \in \{0,\dots,0.2\}$; 200
replications of $T{=}50$ rounds per cell (heatmaps in the supplementary,
Figs.~S1--S2; degenerate $B{=}0$ cells excluded).

\paragraph{H1 --- the flip, with reversed direction.} Past a threshold
$\delta^*(B/N)$, confidence-ranked auditing is \emph{dominated by random}.
We expected a priori that $\delta^*$ decreases as the budget shrinks; the
experiment shows the opposite: \emph{$\delta^*$ increases as $B/N$
shrinks---tight budgets are protected by their own scarcity (they audit only
the extreme low-confidence tail, which stays informative), generous budgets
flip first (they dip into the poisoned mid-ranks)}. Located values:
$\delta^* = \ndstarBNsmall, \ndstarBNfive, \ndstarBNten, \ndstarBNlarge$ at
$B/N = 0.02, 0.05, 0.1, 0.2$; relatively insensitive to
$\rho, \varepsilon, N, F$ in our sweeps
(overconfident-error mass piles at the \emph{high}-confidence end, so the
extreme tail degrades last). On the measured fleet's matched-marginals
background $\delta^*(0.1) \approx \ndstarMatchedGsmLo$
(Fig.~\ref{fig:money}); the flip heatmap is in the supplementary (Fig.~S1).

\paragraph{H2 --- correlation-aware transfer.} The advantage of
\textsf{diversity\_bayes} over random grows with $\rho$ (\nhTwoAdvRhoZero{}
at $\rho{=}0$ to \nhTwoAdvRhoHi{} risk units at $\rho{=}0.8$;
$\delta{=}0.5$, $B/N{=}0.1$). The advantage is robust to the correlation
model (a Beta log-odds shock reproduces the trend) and to the transfer weight
($w \in \{\rho,\rho^2,\rho/2\}$: \nwAdvRho/\nwAdvRhoSq/\nwAdvRhoHalf),
accumulates over rounds, and collapses without persistence (\nhTwoPhiZero{}
at $\varphi{=}0$ vs.\ \nhTwoPhiOne{} at $\varphi{=}1$)---full sweeps in the
supplementary. It needs a not-too-noisy verifier \emph{or} strong
correlation; the $\varepsilon$ crossover is quantified in
\S\ref{sec:limitations}(a).\footnote{Clustering pushes errors beyond any
fixed budget: $\mathbb E[(n_{\mathrm{err}} - B)^+]$ rises \nclusterLo{}$\to$%
\nclusterHi{} as $\rho{=}0{\to}0.8$ ($N{=}100$, $B{=}10$).}

\paragraph{H3 --- vacuous oversight.} With $R_{\mathrm{none}} = \nvacRnone$,
\nvacNa{} of 44 cells are vacuous at $\tau{=}\nvacTau$ (heatmap: Fig.~S2,
supplementary): at $B/N{=}0.02$, oversight in this model is rubber-stamping
almost regardless of $\delta$. Correlation \emph{rescues} high-$\delta$ cells (a
companion $(\delta,\rho)$ sweep at $B/N{=}0.05$ leaves \nvacNb{} vacuous):
with useless confidence, transfer is the only signal.

\section{Measuring the Premises (E2) and Replaying Policies (E3)}
\label{sec:measurement}

\paragraph{Protocol.} Six models---Qwen3 \{0.6B, 4B, 8B\} (one family),
Mistral-7B-Instruct-v0.3, Phi-4-mini-instruct, and hosted \mbox{gpt-4o-mini}
as a cross-provider robustness point---answer the same fixed 500-item
manifests of GSM8K and HotpotQA (distractor) at temperature~0 under one
locked single-completion prompt eliciting an answer and a 0--100 confidence.
Errors are exact-match failures (GSM8K numeric; HotpotQA EM, F1 logged);
two temperature-0.7 runs per open-weight model give same-checkpoint anchors
(per-model values and parse failures $\le\nparseFailMax\%$: Table~S1,
supplementary).

\begin{figure}[t]
\centering
\includegraphics[width=.72\textwidth]{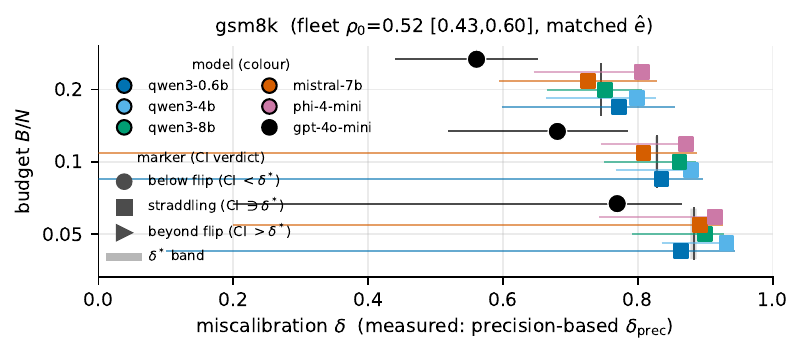}
\caption{Measured fleet on the matched-marginals background (GSM8K; fleet
$\rho_0$, per-model $\hat e$); $\delta_{\mathrm{prec}}$: the
allocation-faithful statistic (\S\ref{sec:measurement}). Colour = model; marker =
CI verdict vs.\ $\delta^*$ (circle below, square straddling, triangle
beyond). Point estimates put \nnBeyondGsm/5 open-weight models beyond the
flip at $B/N{=}0.1$; their CIs straddle $\delta^*$ (long left tails =
flatness of the $\delta\to$precision map at low $\delta$, propagated by the
inversion). gpt-4o-mini (black) sits clearly below the flip (CIs exclude
$\delta^*$ at $B/N \ge 0.1$). Replicates on HotpotQA (Fig.~S3; closest
call $B/N{=}0.05$, $\delta_{\mathrm{prec}}$ \ngptHpqaSmallPrec{} vs.\
$\delta^*{=}\ngptHpqaSmallStar$).}
\label{fig:money}
\end{figure}

\paragraph{Miscalibration is real and extreme (H4).} All five open-weight
models pile verbalized confidence at $\approx$1.0 regardless of correctness
(ECE \neceMin--\neceMax, supplementary); their confidence is near-constant
($\mathrm{Var}(c) \le \nvarCmaxOpen$), so AUROC $\approx 0.5$: \emph{real
small/mid LLMs give operationally useless (near-constant) verbalized
confidence under our locked elicitation protocol, so confidence-ranked
auditing $\approx$ random by construction}.
Because global AUROC is the wrong sufficient statistic for audit value---what
matters is the error composition of the audited tail---we place models by
$\delta_{\mathrm{prec}}$: the error precision of the lowest-confidence $B/N$
tail, inverted through the synthetic $\delta \to \mathrm{precision}$ map at
matched $\hat e$. The six models span the spectrum (Table~S1): the
five open-weight
models have error rates $\hat e \in [\neHatOpenLo, \neHatOpenHi]$ and AUROC
\naurocOpenLo--\naurocOpenHi, with point estimates at or beyond the flip but
CIs straddling $\delta^*$ (inversion sensitivity at high $\hat e$,
Fig.~\ref{fig:money}); the proprietary model has informative
confidence (AUROC
\ngptAurocGsm/\ngptAurocHpqa, positive tail-precision lift) and is correctly
placed below it ($\delta_{\mathrm{prec}} =
\ngptDeltaPrecGsm/\ngptDeltaPrecHpqa$ vs.\ $\delta^* \approx
\ngptDeltaStarGsm$; Fig.~\ref{fig:money}).

\paragraph{Correlation: shared difficulty dominates lineage.} We
expected a priori low cross-family correlation; measured
inter-family latent correlation is \nrhoInterLo--\nrhoInterHi{} (indicator
units \nrhoInterIndLo--\nrhoInterIndHi), overlapping intra-family, and
gpt-4o-mini correlates at \nrhoGptLo--\nrhoGptHi{} with the open-weight
five: the shared factor spans providers. A difficulty-controlled
decomposition (partial correlation given leave-out item difficulty,
item bootstrap) finds a lineage excess $\Delta\rho = \ndRhoGsm$
[\ndRhoGsmLo, \ndRhoGsmHi] (GSM8K) and \ndRhoHpqa{} [\ndRhoHpqaLo,
\ndRhoHpqaHi] (HotpotQA)---both CIs include zero: given item difficulty,
lineage adds no measurable correlation, while a fleet-wide residual remains;
in item-response-theory terms \cite{embretson2000}, shared difficulty
explains co-failure better than ancestry. The two-level copula
(\S\ref{sec:model}) with measured $\rho_0 =
\nfleetRhoZeroGsm/\nfleetRhoZeroHpqa$ (CIs in Fig.~\ref{fig:money}) and
$\rho_f \approx \rho_0$ is the amended model.
H2's mechanism generalizes---transfer pays when persistent errors are
correlated, fleet-wide; the lineage claim softens to ``correlation is partially
predictable from fleet composition.'' Within Qwen3, correlation grades by
size adjacency (4B$\leftrightarrow$8B latent \nrhoAdjGsm/\nrhoAdjHpqa{} vs.\
0.6B$\leftrightarrow$\{4B,8B\} \nrhoSmallLo--\nrhoSmallHi): ``family'' is
not binary.

\paragraph{E3: replay on real traces.} We replay the policies over a
15-agent fleet (5 models $\times$ 3 instances); each round is one benchmark
item, so within-round correlation is as measured. Budgets are
$B \in \{1,2,3\}$ audits on $N{=}15$ (realized fractions 6.7/13.3/20\%).
The oracle uses perfect error knowledge ($1-E$): an upper bound, not a
deployable policy. The E1 ordering survives (replay figures: Figs.~S4--S5,
supplementary): oracle
$<$ diversity-Bayes $<$ conf-ranked $\approx$ random on both benchmarks
(at $B{=}2$, GSM8K: \neThreeOracleGsm{} $<$ \neThreeDivGsm{} $<$
\neThreeConfGsm{} $\approx$ \neThreeRandomGsm). All four
pre-specified paired contrasts are significant on both benchmarks
(Holm-adjusted $p$: random$-$div \nholmRandDivGsmP{}/\nholmRandDivHpqaP{},
conf$-$div \nholmConfDivGsmP{}/\nholmConfDivHpqaP{}, at the 2000-sample
bootstrap floor), with effect sizes \neThreeRandDivGsmMean{} and
\neThreeConfDivGsmMean{} risk units on GSM8K (modest, a floor effect at the
fleet's extreme $\hat e$) and larger \neThreeRandDivHpqaMean{} /
\neThreeConfDivHpqaMean{} on HotpotQA. The structure arbiter confirms the
a-priori prediction: with $\Delta\rho \approx 0$, family $\approx$
tier $\approx$ pooled, capturing all the value.

\section{Limitations}\label{sec:limitations}

\begin{enumerate}
\renewcommand{\labelenumi}{(\alph{enumi})}
\setlength{\itemsep}{0pt}\setlength{\parskip}{0pt}
\item The transfer advantage requires a not-too-noisy verifier \emph{or}
strong correlation: at $\rho{=}0.5$ diversity-Bayes cedes to conf-ranked at
$\varepsilon{=}0.2$, but resists at $\rho{=}0.8$ (crossover values:
supplementary).
\item It requires persistent error profiles (collapses at $\varphi{=}0$), so
fleets under frequent fine-tuning have $\varphi<1$ and an attenuated
advantage.
\item $\hat\rho$ is measured on \emph{aligned} items---an upper bound for
fleets whose agents process different inputs---and is reported in latent
units conditional on the Gaussian copula, extrapolated at high $\hat e$ (so
indicator-unit values are given alongside); the decomposition says which part
(shared difficulty vs.\ lineage residual) transfers.
\item $\hat\delta$ comes from one locked elicitation protocol
(single-completion confidence) and the direct-answer regime (no
CoT); others---separate turn, logits, sampling variance---may recover
variance, and with CoT error rates drop and confidence may change:
a lower-capability operating point, not anomalous.
\item At very high $\delta$ a flip is near-definitional, so
the contribution is \emph{locating} $\delta^*(B/N)$ and its reversed
direction; relatedly, ``AUROC $=0.5$ is the flip boundary'' holds because
measured confidence is near-\emph{constant} (ranking $=$ jitter
$\Rightarrow$ conf-ranked $\equiv$ random), not because the synthetic flip
sits at 0.5 (there it maps to AUROC $\approx \naurocAtFlip$).
\item E3 uses a small fleet ($N{=}15$) and is preliminary.
\item gpt-4o-mini is a single proprietary robustness point, not a sample of
hosted models, and lacks same-checkpoint anchors (hosted temperature-0 is
not deterministic); parse failures are concentrated in one model,
with a declared, non-differential imputation policy (rate and policy:
supplementary).
\end{enumerate}

\section{Conclusion}\label{sec:conclusion}

Trusting self-reported confidence has a failure mode: past $\delta^*(B/N)$
---which rises as budgets shrink---confidence-ranked auditing is worse than
random, where the five measured open-weight models' point estimates sit.
Correlation, driven by fleet-wide shared difficulty not lineage, is both a
threat (clustering errors beyond any budget) and the only resource when
confidence is useless; pooled transfer needs no lineage map. The vacuity
criterion gives a quantitative, model-level proxy for ``effective human
oversight'': below it, no studied policy beats rubber-stamping. \emph{Future work}: a multi-provider
survey, the strategic case (AI Control), sequential within-round auditing,
and richer elicitation.

\begin{credits}
% Acknowledgments omitted for the double-blind submission (was a scrubbed
% placeholder, and it cost the line that pushed the disclosure to p7). Re-add
% camera-ready: \subsubsection{\ackname} <funding/thanks>.
\subsubsection{\discintname} The authors have no competing interests to
declare.
\end{credits}

\bibliographystyle{splncs04}
\bibliography{references}

\newpage
\begin{center}
{\Large Appendix}
\end{center}

This appendix collects illustrative material for the main paper.
Every load-bearing number is reported in the main text; the figures and
tables here only illustrate. Figures and tables are numbered S1, S2,
\dots; references to the main paper are given by description.

\section{Synthetic phase diagrams (E1)}\label{app:phase}

\begin{figure}[H]
\centering
\includegraphics[width=.6\textwidth]{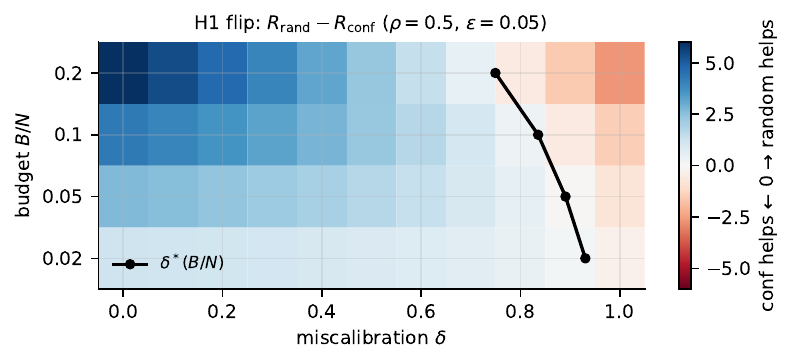}
\caption{H1 flip (synthetic; $N{=}100$, $F{=}4$; theory background,
$e_i \sim U(0.05,0.30)$): $R_{\mathrm{random}} - R_{\mathrm{conf}}$ over
$(\delta, B/N)$; the black curve is the flip threshold $\delta^*(B/N)$,
\ndstarMonotone{} in $B/N$. The located values, the reversed direction, and
the matched-fleet placement are stated in full in the main paper (\S4).}
\label{fig:flip}
\end{figure}

\begin{figure}[H]
\centering
\includegraphics[width=.6\textwidth]{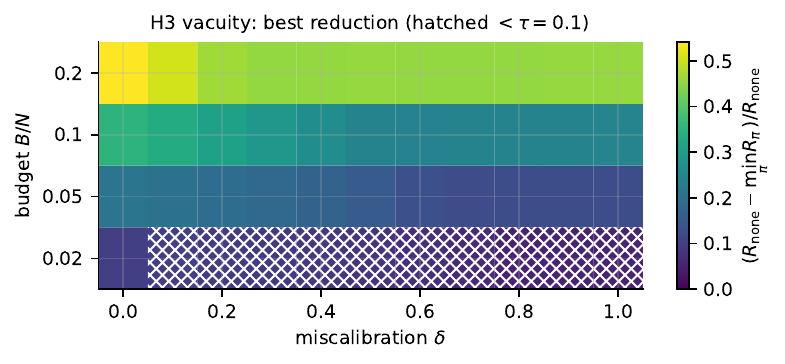}
\caption{H3 vacuity (synthetic; same background as Fig.~\ref{fig:flip}): best
risk reduction over the \emph{deployable} policies considered,
$(R_{\mathrm{none}}-\min_\pi R_\pi)/R_{\mathrm{none}}$; hatched cells are
vacuous ($<\tau{=}\nvacTau$). \nvacNa{} of 44 cells are vacuous; at
$B/N{=}0.02$ mandated oversight is rubber-stamping almost regardless of
$\delta$ (main paper, \S4, H3).}
\label{fig:vacuity}
\end{figure}

\section{Per-model measurements (E2)}\label{app:table}

\begin{table}[H]
\centering
\caption{Per-model measurements: error rate $\hat e$, AUROC of confidence
for correctness, precision-based miscalibration $\delta_{\mathrm{prec}}$ at
$B/N{=}0.1$, and expected calibration error (ECE). 500 items per benchmark,
temperature 0, locked single-completion elicitation.}
\label{tab:e2}
\setlength{\tabcolsep}{4.2pt}
\begin{tabular}{l cccc cccc}
\toprule
& \multicolumn{4}{c}{GSM8K} & \multicolumn{4}{c}{HotpotQA} \\
\cmidrule(lr){2-5}\cmidrule(lr){6-9}
Model & $\hat e$ & AUROC & $\delta_{\mathrm{prec}}$ & ECE
      & $\hat e$ & AUROC & $\delta_{\mathrm{prec}}$ & ECE \\
\midrule
\neTwoRowsBoth
\bottomrule
\end{tabular}
\end{table}

\section{Measured fleet on HotpotQA}\label{app:money_hpqa}

\begin{figure}[H]
\centering
\includegraphics[width=.6\textwidth]{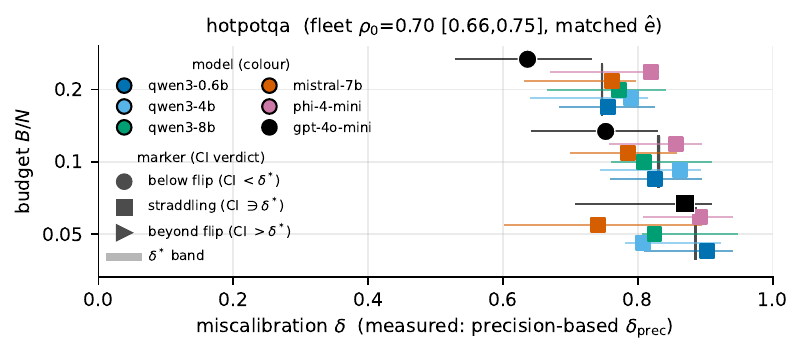}
\caption{HotpotQA counterpart of the GSM8K money diagram (main paper,
Fig.~1; same axes, markers, and legend). \nnBeyondHpqa/5 open-weight models
are beyond the flip at $B/N{=}0.1$; gpt-4o-mini sits below it, with its
closest call at $B/N{=}0.05$ ($\delta_{\mathrm{prec}}$ \ngptHpqaSmallPrec{}
vs.\ $\delta^*{=}\ngptHpqaSmallStar$).}
\label{fig:money_hpqa}
\end{figure}

\section{E3 replay on real traces}\label{app:replay}

The E3 replay (main paper, \S5) carries all its numbers inline---the
ordering, the four Holm-adjusted contrasts, and the effect sizes; these two
panels illustrate it on both benchmarks.

\begin{figure}[H]
\centering
\includegraphics[width=.6\textwidth]{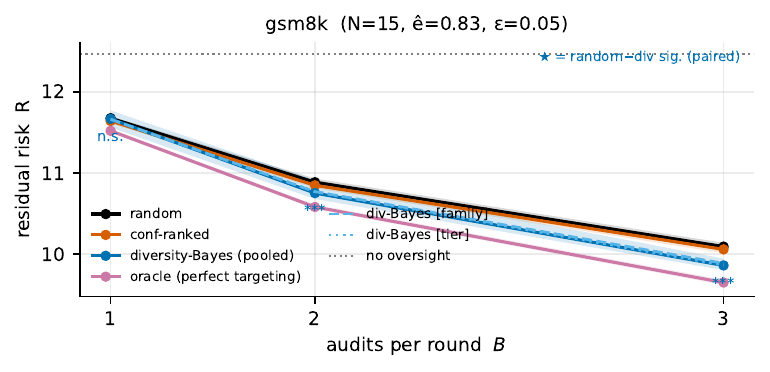}
\caption{E3 replay on GSM8K traces (15 agents, $\varepsilon{=}0.05$):
residual risk vs.\ budget; stars mark the pre-specified random$-$diversity
contrast; oracle = perfect error knowledge (reference); diversity-Bayes under
family (dashed) / tier (dotted) / pooled coincides ($\Delta\rho \approx 0$).}
\label{fig:replay_gsm}
\end{figure}

\begin{figure}[H]
\centering
\includegraphics[width=.6\textwidth]{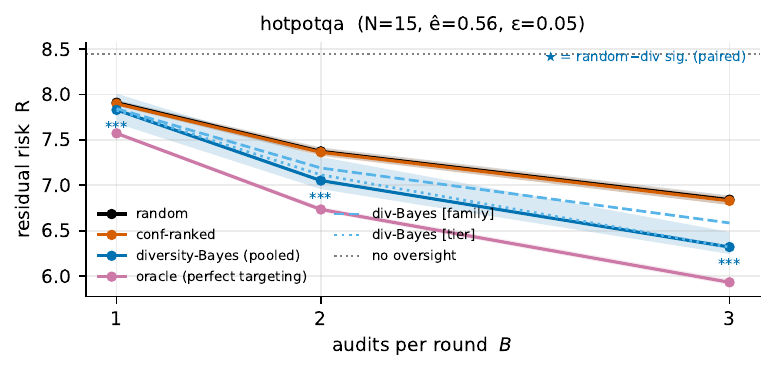}
\caption{HotpotQA counterpart of Fig.~\ref{fig:replay_gsm} (same axes and
policies). The ordering oracle $<$ diversity-Bayes $<$ conf-ranked $\approx$
random holds, and family/tier/pooled groupings coincide. The contrast effect
sizes (\neThreeRandDivHpqaMean{} / \neThreeConfDivHpqaMean{} risk units) are
larger than on GSM8K, as reported in the main text (\S5).}
\label{fig:replay_hpqa}
\end{figure}

\section{H2 in the synthetic model; robustness to the copula choice}
\label{app:sensitivity}

E1 grid cells with $N{=}20$, $B/N{=}0.02$ realize an integer budget
$B{=}0$ (all policies coincide); they are excluded from every figure and
number, and the exclusion is asserted in the figure code.

The H2 advantage accumulates over rounds (\nhTwoTten{} at $T{=}10$,
\nhTwoTfifty{} at $T{=}50$, \nhTwoTtwohundred{} at $T{=}200$; $\rho{=}0.8$)
and depends critically on persistence: \nhTwoPhiZero{} at $\varphi{=}0$
(per-round shocks) vs.\ \nhTwoPhiOne{} at $\varphi{=}1$---the transfer
mechanism is learning persistent family error profiles from audits. (The
persistence collapse is also reported inline in the main paper, \S4.)

\begin{figure}[H]
\centering
\includegraphics[width=.72\textwidth]{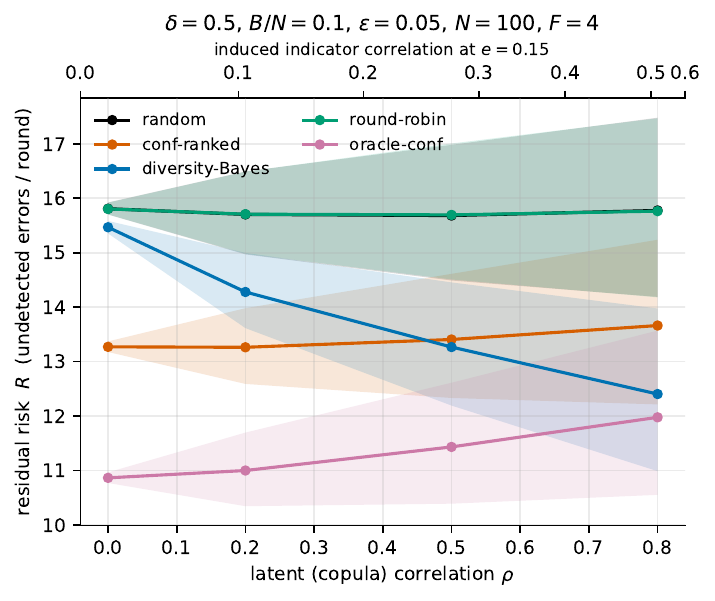}
\caption{H2 (synthetic, amended model): residual risk vs.\ latent $\rho$ at
$\delta{=}0.5$, $B/N{=}0.1$, $\varepsilon{=}0.05$. diversity-Bayes pulls
away from random as $\rho$ grows; conf-ranked and the oracle drift slightly
upward because clustered errors exceed the fixed budget more often (main
paper, \S4).}
\label{fig:h2}
\end{figure}

\begin{figure}[H]
\centering
\includegraphics[width=.72\textwidth]{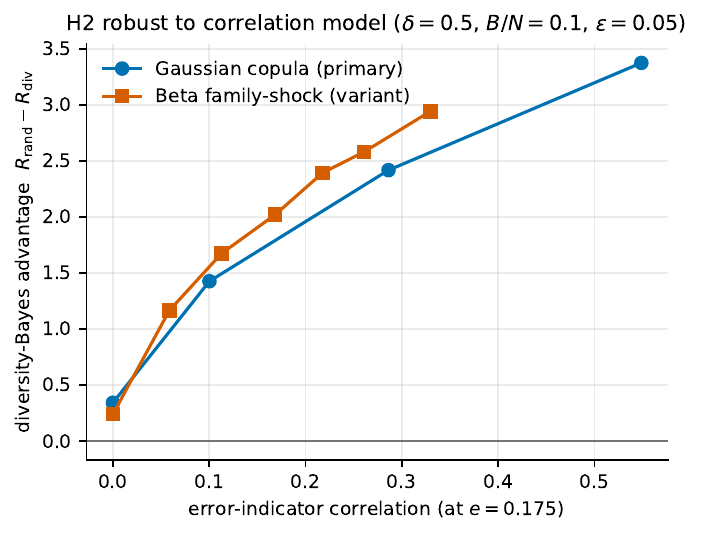}
\caption{Sensitivity: a marginal-preserving Beta log-odds family-shock error
model reproduces the H2 trend (advantage up to \nbetaShockAdvHi{} at the
strongest shock) and the H1 thresholds, so neither finding is an artifact of
the Gaussian copula. Transfer-weight choice $w \in \{\rho, \rho^2, \rho/2\}$
(\nwAdvRho/\nwAdvRhoSq/\nwAdvRhoHalf{} at $\rho{=}0.8$) changes convergence
speed, not conclusions.}
\label{fig:betashock}
\end{figure}

\section{Reliability diagrams}\label{app:reliability}

\begin{figure}[H]
\centering
\includegraphics[width=.95\textwidth]{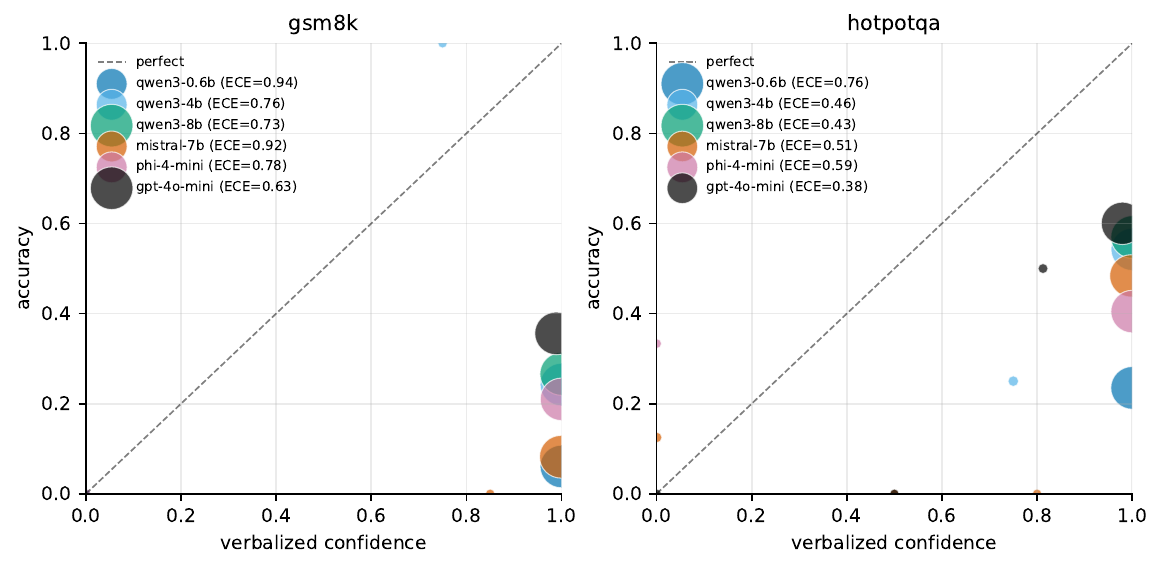}
\caption{Reliability diagrams for the six models. Open-weight confidence
piles at $\approx$1.0 with accuracy far below the diagonal (ECE
\neceMin--\neceMax); gpt-4o-mini (black) is the only model whose confidence
variation is error-correlated.}
\label{fig:reliability}
\end{figure}

\section{Confidence parse failures and imputation}\label{app:parse}

Confidence parse failures (no numeric confidence in the completion) are at
most \nparseFailMax\% per trace and concentrate in Qwen3-0.6B; they are zero
for four of the six models. Two consumers handle them differently, by
design: analysis statistics (AUROC, $\mathrm{Var}(c)$,
$\delta_{\mathrm{prec}}$) \emph{drop} unparsed records, while the E3 replay
\emph{imputes} maximum confidence (1.0), so conf-ranked audits those outputs
last. The asymmetry is declared, affects \nreplayImputedPct\% of replay
cells, and is non-differential across the conclusions: the AUROC
$\approx 0.5$ finding is shared by four models with zero parse failures.

\section{Limitation support values}\label{app:limvals}

Quantifying values for the boundary statements in the main paper's
Limitations (\S6); each limitation states its boundary in full, with the
supporting numbers here.
\begin{itemize}
\item[(a)] Verifier-noise crossover at $\varepsilon{=}0.2$: at $\rho{=}0.5$
diversity-Bayes \nepsCrossDivLo{} vs.\ conf-ranked \nepsCrossConfLo{} (cedes);
at $\rho{=}0.8$, \nepsCrossDivHi{} vs.\ \nepsCrossConfHi{} (resists).
\item[(g)] Parse-failure rate $\le\nparseFailMax\%$, concentrated in
Qwen3-0.6B (details and imputation policy in \S\ref{app:parse}).
\end{itemize}
\end{document}